\documentclass{article}
\usepackage{spconf,amsmath,epsfig,amssymb,latexsym,multirow}
\usepackage{fancyhdr}

\usepackage{color}

\begin{document}\sloppy

\def\x{{\mathbf x}}
\def\L{{\cal L}}

\title{Fine-Grained Expression Manipulation via structured latent Space}
%
\name{Junshu Tang$^{1}$, Zhiwen Shao$^{\ast1}$, and Lizhuang Ma$^{\ast1,2}$\thanks{$^\ast$ Corresponding authors. This work is supported by the National Natural Science Foundation of China (No. 61972157). This is also partially supported by the Wu Wen Jun Honorary Doctoral Scholarship, AI Institute, Shanghai Jiao Tong University.}}
\address{$^{1}$Department of Computer Science and Engineering, Shanghai Jiao Tong University, China\\
$^{2}$School of Computer Science and Technology, East China Normal University, China\\
\{tangjs,shaozhiwen\}@sjtu.edu.cn, ma-lz@cs.sjtu.edu.cn.}

\maketitle

\thispagestyle{fancy}
\fancyhead{}
\lhead{}
\lfoot{978-1-7281-1331-9/20/\$31.00~\copyright 2020 IEEE}
\cfoot{}
\rfoot{}

\begin{abstract}
Fine-grained facial expression manipulation is a challenging problem, as fine-grained expression details are difficult to be captured. Most existing expression manipulation methods resort to discrete expression labels, which mainly edit global expressions and ignore the manipulation of fine details. To tackle this limitation, we propose an end-to-end expression-guided generative adversarial network (EGGAN), which utilizes structured latent codes and continuous expression labels as input to generate images with expected expressions. Specifically, we adopt an adversarial autoencoder to map a source image into a structured latent space. Then, given the source latent code and the target expression label, we employ a conditional GAN to generate a new image with the target expression. Moreover, we introduce a perceptual loss and a multi-scale structural similarity loss to preserve identity and global shape during generation. Extensive experiments show that our method can manipulate fine-grained expressions, and generate continuous intermediate expressions between source and target expressions.


\end{abstract}
\section{Introduction}
\label{sec:intro}
Facial expression manipulation is an important task which aims to manipulate the facial expression of an image without changing its identity. It has a wide application prospect in digital entertainment, film production, and other fields. The goal of this task is to translate a source image into a new image with the target expression. However, in literature manipulating fine-grained expression details is still a challenging problem, since fine details are difficult to be captured.

To generate photo-realistic images, generative models such as variational autoencoders (VAEs)~\cite{kingma2013auto} and generative adversarial networks (GANs)~\cite{goodfellow2014generative} are extensively used in recent years and have achieved great success. Conditional GAN (cGAN)~\cite{mirza2014conditional} takes advantage of a conditional variable to control the generated data in a supervised way. IcGAN~\cite{perarnau2016invertible} further maps input images into a high-level latent space and can re-generate the original images with complex variations.
Since paired data are often hard to collect, some methods like UNIT~\cite{liu2017unsupervised} map a group of corresponding images in different domains to the same latent representation in a shared latent space for unsupervised learning.

By treating the expression as a kind of facial attribute, most existing methods~\cite{liu2017unsupervised,Choi_2018_CVPR,he2019attgan,chen2019Homomorphic} group expressions with other attributes like hairstyle, gender and age, and represent these discrete labels as a one-hot vector to control the attribute manipulation. However, it is incomplete to use only a single number or a discrete label to describe an expression that consists of local muscle actions. Besides, different people show different appearances for the same expression. For instance, in a smiling expression, some people only raise up the left mouth corner while other people raise up the right mouth corner. To describe fine-grained facial expressions, Ekman et al.~\cite{ekman1997face} developed a Facial Action Coding System (FACS) which defines anatomically facial action units (AUs). Each AU is a basic action of a single muscle or a group of muscles. By using AUs with continuously varied intensities, we can code any expression with fine-grained details.

\begin{figure}
    \centering
    \includegraphics[width=1\linewidth]{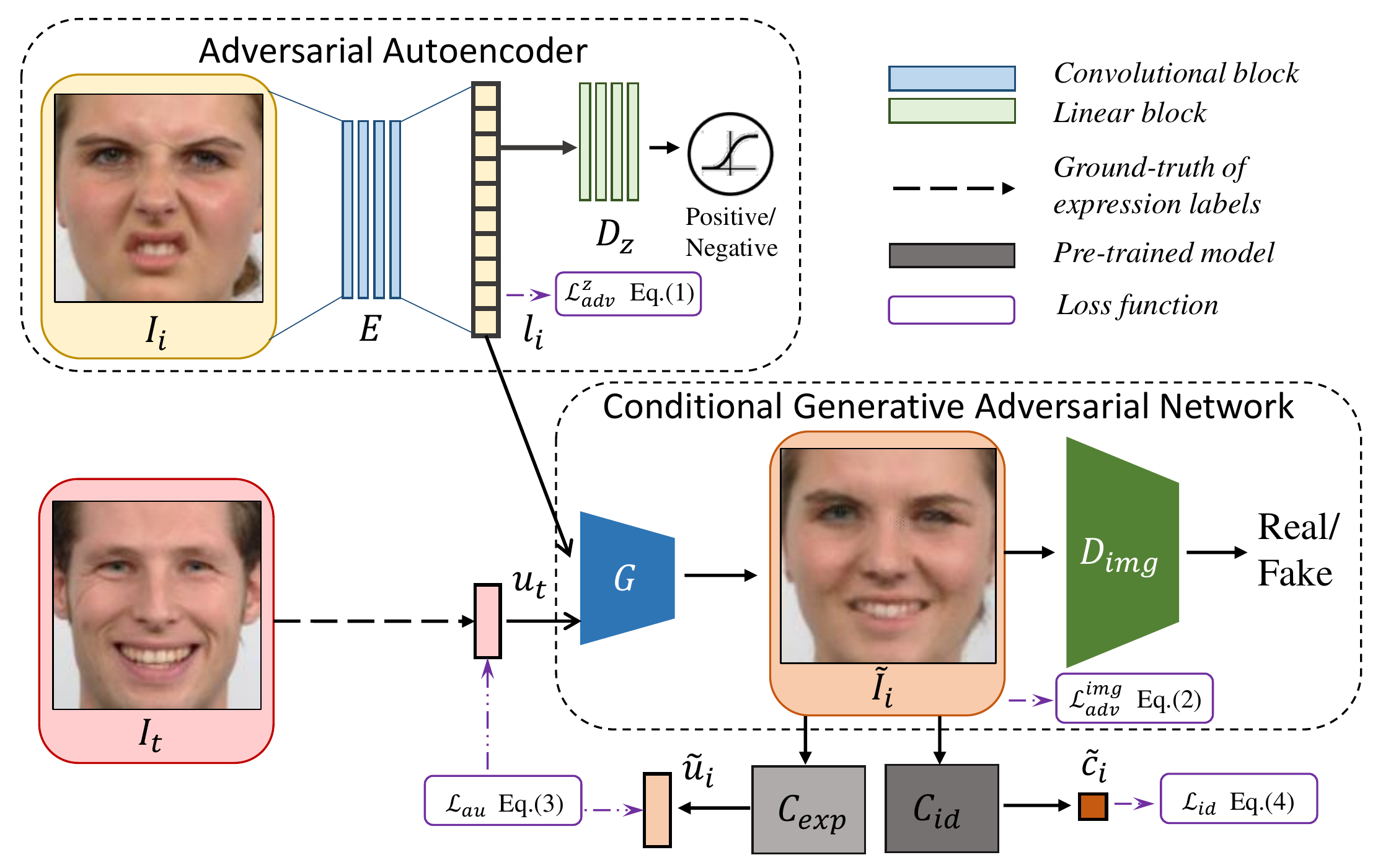}
    \caption{The architecture of our framework, which consists of an AAE and a cGAN. Given a source face image $I_{i}$ and a target AU intensity vector $u_{t}$, our framework generates a new image $\tilde{I}_{i}$ which contains the target expression and other preserved information including identity and global facial shape.}
    \label{fig:basenetwork}
    \vspace{-0.1in}
\end{figure}

In this paper, we propose a novel Expression-Guided GAN (EGGAN) for fine-grained expression manipulation. Fig.~\ref{fig:basenetwork} illustrates the architecture of our framework. In particular, we use an adversarial autoencoder (AAE)~\cite{makhzani2015adversarial} to map the input source image into a structured latent space. Then, given the continuous AU intensities of a target image and a structured latent code containing rich information of the source image, we employ a cGAN to generate a new image with the target expression. 
Moreover, we introduce a perceptual loss and a multi-scale structural similarity (MS-SSIM) loss~\cite{wang2004image} to preserve identity and structural information in the generated images. Our framework is trainable end-to-end, and can manipulate fine-grained facial expressions in a continuous interpolation space. 

The main contributions of this paper are threefold: (i) We propose to map the input image into a structured latent space, which is beneficial for fine-grained expression manipulation. 
(ii) We introduce a perceptual loss and a MS-SSIM loss, which contributes to preserving identity and global facial shape in the generated images, respectively.
(iii) Experimental results demonstrate that our method can manipulate fine-grained expression details, and can generate continuous interpolation results between source and target expressions.

\vspace{-0.08in}
\section{Related Work}

\noindent\textbf{Conditional Generative Adversarial Network.} 
GANs~\cite{goodfellow2014generative} have been extensively used, which are performed as a two-player minimax game. cGAN~\cite{mirza2014conditional} is further proposed to control the distribution of generated data by using additional condition information. Recently, there are many works employing the cGAN for image synthesis, in which the conditional code usually contains rich information of images. DRIT~\cite{DRIT} disentangles images into a content latent space and a style latent space for style transfer. StyleGAN~\cite{Karras_2019_CVPR} applies a nonlinear transformation on latent codes to strengthen the control of different image styles.
Different from these methods, our approach maps input images to a structured latent space, which utilizes continuous AU intensities and structured latent codes.

\noindent\textbf{Facial Attribute Manipulation.}
Facial attribute manipulation is a popular topic in computer vision field.
Recent works mainly use the GAN to generate photo-realistic results with expected attributes. Most of these methods group discrete attributes as one-hot vectors, and can edit an attribute from non-occurrence to occurrence by changing its value from $0$ to $1$. StarGAN~\cite{Choi_2018_CVPR} and AttGAN~\cite{he2019attgan} translate images to different domains by inputting both the given image and target domain information to the generator. Since facial expression is a complex attribute with many local details, it is hard to distinguish all the expressions using limited domains, especially for some micro-expressions. Instead of using discrete expression labels, GANimation~\cite{pumarola2018ganimation} controls fine-grained transformation strength of two domains using continuous AU intensity labels. 
In contrast, we employ a more simplified training framework for fine-grained expression manipulation.

\vspace{-0.08in}
\section{Method}
\vspace{-0.15in}
\subsection{Overview}
Given a source image $I_{i}\in\mathbb{R}^{H\times W\times3}$ and a target AU intensity vector $u_{t}=[u^{1}_{t},u^{2}_{t},...,u^{d}_{t}]$, $d$ is the number of AUs, our goal is to generate a new image which has the target expression while preserving the original information about identity and global facial shape. Besides, we aim to obtain intermediate results after linear interpolation between source and target AU intensity vectors. 
Fig.~\ref{fig:basenetwork} shows the architecture of our EGGAN framework. It consists of an AAE and a cGAN, in which the former contains an encoder $E$ and a latent discriminator $D_{z}$, the latter contains a generator $G$ and an image discriminator $D_{img}$. 

Specifically, the encoder $E$ encodes source image $I_{i}$ into a structured latent code $l_{i}$. The latent discriminator $D_{z}$ distinguishes the code $l_{i}$ from a random vector $z$ in a Gaussian distribution. The generator $G$ further employs $l_{i}$ and $u_{t}$ to generate a new image $\tilde{I}_{i}$ with the expected target expression. 
The adversarial learning in AAE aims at limiting the diversity of encoded data from source images and encouraging the images generated by interpolating between arbitrary identity representations not to deviate from the face manifold. The key of our work lies in mapping real images into a structured latent space, which contains structural information including identity and global facial shape. 




\vspace{-0.1in}
\subsection{Network Modules}

\noindent\textbf{Adversarial Autoencoder.} 
In the AAE, given an input image $I_{i}$, $E$ outputs a latent code $l_{i} = E(I_{i})\thicksim q_{z}$. We apply a discriminator $D_{z}$ to distinguish between $l_{i}$ and a randomly sampled vector $z\thicksim p_{z}$, where $p_{z}= N(\mu_{z},\sigma_{z})$ is a Gaussian distribution as the prior distribution. AAE can encode images in a local Euclidean space with structural information to ensure the interpolation of encoded data is still on a manifold. We employ a latent adversarial loss $\mathcal{L}_{adv}^{z}$ to learn the structured latent mapping:
\begin{equation}
    \begin{aligned}
        \mathcal{L}_{adv}^{z}&=\mathbb{E}_{z\sim p_{z}}[\log D_{z}(z)]\\&+\mathbb{E}_{I_{i}\sim p_{data}}[\log(1-D_{z}(E(I_{i})))],
    \end{aligned}
\end{equation}
where $I_{i}$ is sampled from the image domain $p_{data}$, and $z$ follows the Gaussian distribution $p_{z}$. $\mathcal{L}_{adv}^{z}$ encourages $l_{i}$ to be indistinguishable from $z$.  

\noindent\textbf{Conditional Generative Adversarial Network.}
In the cGAN, we use the target AU intensity vector $u_t$ as the conditional variable to generate an image conditioned on the expected expression. The image discriminator $D_{img}$ is trained to make the generated image $\tilde{I}_i=G(l_i|u_t)$ indistinguishable from the real image. We use a recently proposed method WGAN-GP~\cite{gulrajani2017improved} to adversarially train our cGAN, which applies a gradient penalty to enforce the Lipschitz~\cite{arjovsky2017wasserstein} 
constraint and avoid weight clipping in traditional GANs. The image adversarial loss is defined as:
\begin{equation}
    \begin{aligned}
    \mathcal{L}_{adv}^{img}=&\mathbb{E}_{I_{i}\sim p_{data}}[\log D_{img}(I_{i})]+\\&\mathbb{E}_{l_{i}\sim p_{l}}[\log(1-D_{img}(G(l_i|u_{t})))]+\\
    &\lambda_{gp}\mathbb{E}_{I_{i}\sim p_{data}}[(\|\nabla _{I_{i}} D_{img}(I_{i})\|_{2}-1)^2],
    \label{con:imgadvloss}
    \end{aligned}
\end{equation}
where the last item is a soft version of gradient penalty for random samples $I_{i}\sim p_{data}$, and $\lambda_{gp}$ controls the importance of the gradient penalty term.

\vspace{-0.1in}
\subsection{Attribute Constraint}

The goal of our method is to synthesize an image $\tilde{I}_i$ with expected fine-grained expression and original identities. We adopt an expression classifier $C_{exp}$ and an identity classifier $C_{id}$ with the same architecture, in which the former classifies each AU intensity from multiple levels, and the latter ensures the generated images to preserve the identity information. 
The architecture of these two classifiers are inspired by the traditional VGG19~\cite{simonyan2014very}. On account of the high semantic of identity, we pre-train $C_{id}$ independently using training images $I_i$ and their identity class labels $c_i$. 

\noindent\textbf{Expression Constraint.}
Since the expression is decomposed into a set of intensity values of AUs, we regard expression classification as a multi-level intensity classification problem. In particular, we discretize a continuous intensity value $u_t^j$ as $\lfloor u_t^j \rceil \in \{0,\cdots,m\}$, where $\lfloor \cdot \rceil$ denotes the operation of rounding a number to the nearest integer, and $m=5$ is the maximum intensity level.
The AU intensity classification loss for $\tilde{I}_i$ is defined as
\begin{equation}
    \begin{aligned}
        \mathcal{L}_{au}=
        -\frac{1}{d}\sum^{d}_{j=1}\sum^{m}_{q=0}\mathbf{1}_{q=\lfloor u^j_{t} \rceil}\log(C_{exp}^{(j,q)}(\tilde{I}_i)),
    \end{aligned}
\end{equation}
where $C_{exp}$ outputs a $d(m+1)$-dimensional vector which predicts the probability of each AU intensity, $C_{exp}^{(i,q)}(\tilde{I}_i)$ denotes the $q$-th value of $i$-th AU output by $C_{exp}$, and $\mathbf{1}_{[.]}$ denotes the indicator function.
$\mathcal{L}_{au}$ encourages $C_{exp}$ to predict the probability of $1$ for the index of $\lfloor u^j_{t} \rceil$ while predicting $0$ for other indexes.

\noindent\textbf{Identity Constraint.}
To preserve identity information for $\tilde{I}_i$, we use an identity classification loss:
\begin{equation}
    \begin{aligned}
        \mathcal{L}_{id} = -\sum^{n}_{k=1} \mathbf{1}_{k=c_i} \log(C^{(k)}_{id}(\tilde{I}_i)),
    \end{aligned}
\end{equation}
where $n$ is the number of identity classes. Moreover, to suppress low resolution and artifacts in the generated images, we additionally introduce a perceptual loss $\mathcal{L}_{per}$~\cite{10.1007/978-3-319-46475-6_43} to constrain the identity preservation in expression manipulation and self-reconstruction:
\begin{equation}
    \begin{aligned}
        \mathcal{L}_{per}^{\phi,3} =\frac{1}{C_{3}H_{3}W_{3}} [&\|\phi_{3}(I_{i})-\phi_{3}(\tilde{I}_{i})\|_{2}+\\
        &\|\phi_{3}(I_{i})-\phi_{3}(\hat{I}_{i})\|_{2}],
    \end{aligned}
\end{equation}
where 
$\phi_{3}$ denotes the activation of the third layer in the official pre-trained VGG19 model, and $\hat{I}_{i}=G(E(I_{i})|u_{i})$ is the self-reconstructed image.

\par
\vspace{-0.1in}
\subsection{Full Objective Function}

To ensure the stability of image generation, $G$ should be able to self-reconstruct $I_{i}$. However, L1 norm often results in structural distortion and image blurring. To alleviate these problems, we apply an L1 loss and a MS-SSIM loss~\cite{wang2004image} to constrain self-reconstruction:
\begin{equation}
    \begin{aligned}
        \mathcal{L}_{rec}=\mathbb{E}[\|\hat{I}_{i}-I_{i}\|_{1}] +(1-SSIM(\hat{I}_{i}, I_{i})),
    \end{aligned}
\end{equation}
where $SSIM(\cdot)$ denotes the MS-SSIM loss, which is beneficial for preserving global facial shape.

Combining the losses introduced above, the full objective function is formulated as
\begin{equation}
    \begin{aligned}
        \min_{E,G,C_{exp}}\max_{D_{z},D_{img}} \mathcal{L} &= \mathcal{L}_{adv}^{z} + \mathcal{L}_{adv}^{img} + \lambda_{au}\mathcal{L}_{au} \\&+ \lambda_{id}\mathcal{L}_{id} + \lambda_{per}\mathcal{L}_{per}^{\phi,3} + \lambda_{rec}\mathcal{L}_{rec},
        \label{con:fulllossfunction}
    \end{aligned}
\end{equation}
where $\lambda_{(\cdot)}$ controls the relative importance of each loss term.
\vspace{-0.15in}
\begin{figure*}
    \centering
    \includegraphics[width=1\linewidth]{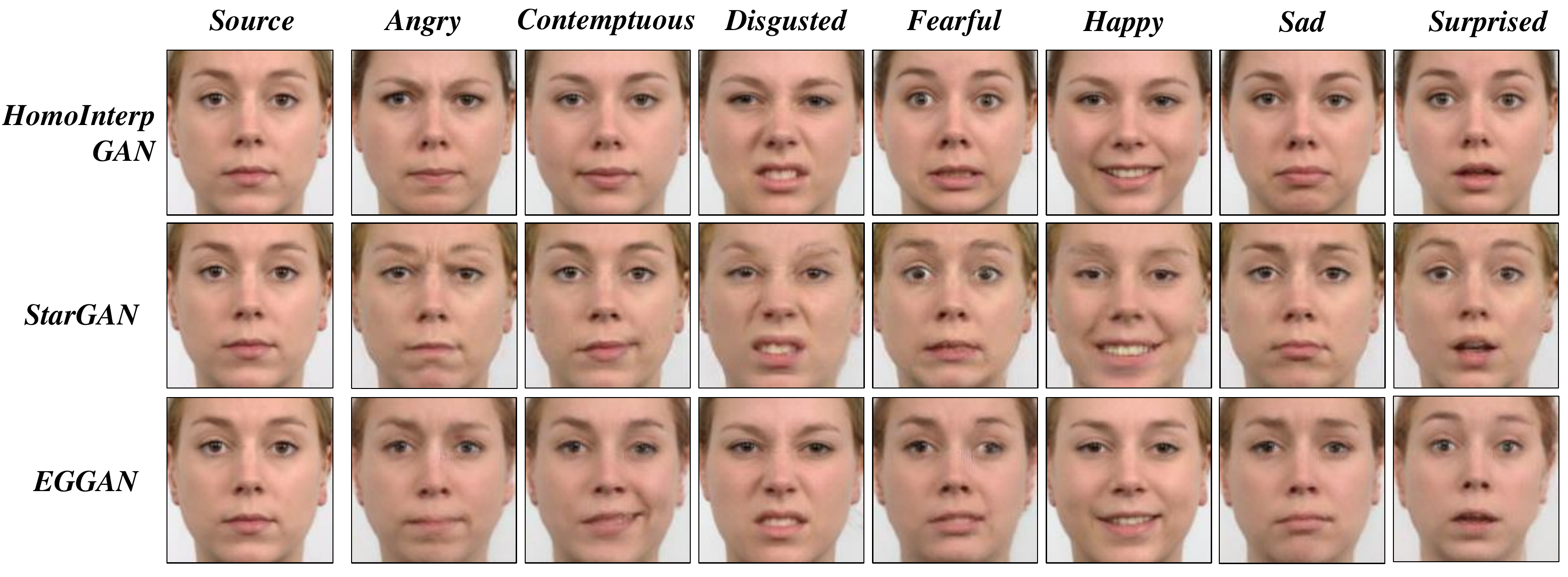}
    \vspace{-8mm}
    \caption{Visual comparison of different expressions for HomoInterpGAN, StarGAN and our EGGAN on the RaFD dataset. The input source image has a neutral expression.}
    \label{fig:compairresult}
\end{figure*}

\begin{table*}[t]
\setlength{\tabcolsep}{1mm}
\begin{center}
\begin{tabular}{|c|c|c|c|c|c|c|c|c|c|c|c|c|c|c|c|c|c|c|c|c|}
\hline
\multicolumn{2}{|c|}{AU} & 1 & 2 & 4 & 5 & 6 & 7 & 9 & 10 & 12 & 14 & 15 & 17 & 20 & 23 & 25 & \multicolumn{1}{c|}{26} & 45 & \bf{Avg} \\ \hline
\multicolumn{1}{|c|}{\multirow{3}{*}{MSE}} & HGAN & 0.56 & 0.86 & 0.88 & 1.68 & 0.23 & 1.26 & 0.60 & \textbf{0.08} & \textbf{0.03} & \textbf{0.11} & 0.17 & \textbf{0.08} & \textbf{0.25} & \textbf{0.01} & \textbf{0.10} & \textbf{0.13} & 0.02 & 0.41\\
\multicolumn{1}{|c|}{}  & SGAN & 0.85 & 1.59 & 0.66 & 1.62 & 0.43 & 1.04 & 0.59 & 0.26 & 0.77 & 0.27 & 0.21 & 0.14 & 0.30 & \textbf{0.01} & 1.09 & 0.39 & \textbf{0.00} & 0.60\\
\multicolumn{1}{|c|}{} & \textbf{EGAN} & \textbf{0.13} & \textbf{0.13} & \textbf{0.49} & \textbf{0.35} & \textbf{0.21} & \textbf{0.49} & \textbf{0.33} & 0.14 & 0.41 & 0.16 & \textbf{0.15} & 0.18 & 0.26 & \textbf{0.01} & 0.51 & 0.14 & \textbf{0.00} & \textbf{0.24}\\ \hline
\multirow{3}{*}{PCC}  & HGAN & 0.64 & 0.68 & \textbf{0.69} & 0.35 & \textbf{0.83} & 0.41 & 0.55 & \textbf{0.83} & \textbf{0.98} & \textbf{0.76} & \textbf{0.71} &\textbf{0.78} &\textbf{0.55}& \textbf{0.32} & \textbf{0.93} & \textbf{0.79} & 0.26 & 0.65 \\ 
 & SGAN & 0.34 & 0.00 & 0.64 & 0.08 & 0.24 & 0.27 & 0.34 & 0.31 & 0.33 & 0.15 & 0.40 & 0.57 & 0.48 & 0.20 & 0.45 & 0.15 & 0.19 & 0.30  \\ 
& \textbf{EGAN} & \textbf{0.92} & \textbf{0.95} & \textbf{0.69} & \textbf{0.85} & 0.75 & \textbf{0.72} & \textbf{0.73} & 0.70 & 0.78 & 0.65 & 0.64 & 0.33 & 0.50 & 0.22 & 0.67 & 0.76 & \textbf{0.34} & \textbf{0.66}\\ \hline
\end{tabular}
\end{center}
\vspace{-3mm}
\caption{Quantitative evaluation of expression manipulation for HomoInterpGAN, StarGAN and our EGGAN. We compute MSE (lower is better) and PCC (higher is better) between 17 AU intensities of target images and generated images. We also report the average results over all AUs (Avg). HomoInterpGAN, StarGAN and EGGAN are shortly written as HGAN, SGAN and EGAN, respectively.}
\vspace{-0.1in}
\label{tab:cacuAU}

\end{table*}

\begin{table}
\setlength{\tabcolsep}{3mm}
\begin{center}
\begin{tabular}{|c|c|c|c|c|}
\hline
Method & HGAN &SGAN&\textbf{EGAN}&Real \\ \hline
Accuracy &76.54&93.32&\textbf{97.04}&99.63\\ \hline
\end{tabular}
\end{center}
\vspace{-3mm}
\caption{Quantitative evaluation of identity preservation for different methods. 
``Real'' contains pairs of real images with the upper-bound results.}
\vspace{-3mm}
\label{tab:cacuID}
\end{table}

\section{Experiments}
\subsection{Implementation Details}

We train our method on RaFD~\cite{langner2010presentation} and MMI~\cite{pantic2005web} datasets. Since RaFD and MMI do not provide AU intensity labels, we use the AU intensity predictor in a popular library OpenFace~\cite{baltrusaitis2018openface} to annotate intensities of 17 AUs (1, 2, 4, 5, 6, 7, 9, 10, 12, 14, 15, 17, 20, 23, 25, 26 and 45) as continuous expression labels. We centrally crop images by face alignment and resize them to $128\times128$.

\begin{figure*}[tbh]
    \centering
    \includegraphics[width=1\linewidth]{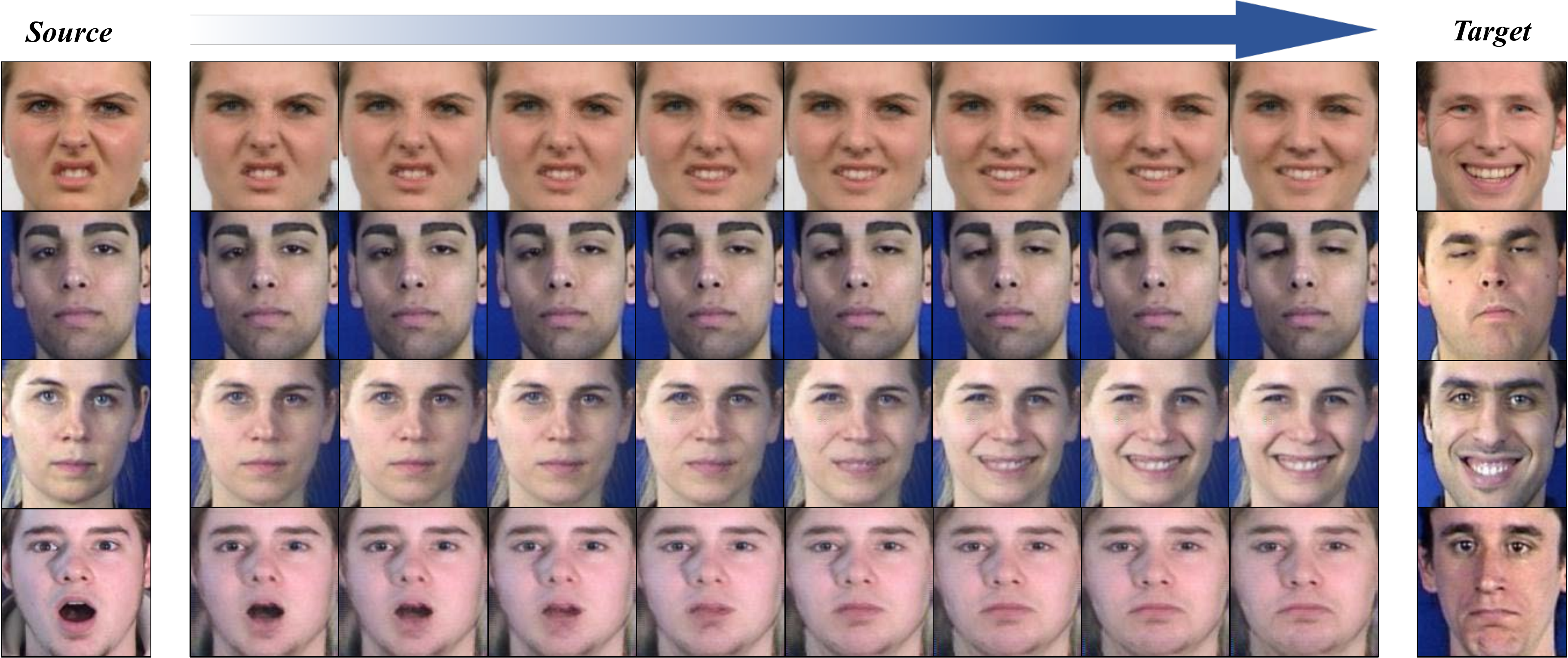}
    \vspace{-6mm}
    \caption{Illustration of fine-grained expression interpolation. The generated images from left to right show expressions from source to target, in which the identity and global facial shape are preserved.}
    \vspace{-6mm}
    \label{fig:interpshow}
\end{figure*}

In our EGGAN framework, the generator $G$ is composed of $6$ up-sampling residual blocks, in which we use Instance Normalization and PReLU as activation. The image discriminator $D_{img}$ 
is composed of $6$ convolution layers with a stride of $2$. The encoder $E$ consists of $5$ convolution layers with a stride of $2$ in $16$, $32$, $64$, $128$ and $256$ channels respectively, followed by a $1$ fully-connected layer. The latent discriminator $D_{z}$ is comprised of $6$ fully connected layer with $256$, $128$, $64$, $32$, $16$ and $1$ channels, respectively. 

We set the hyper-parameters as: $\lambda_{au}=100$, $\lambda_{id}=60$, $\lambda_{per}=20$, $\lambda_{rec}=100$ and $\lambda_{gp}=20$. The adversarial learning in $E$, $G$, $D_z$ and $D_{img}$ employs the Adam solver and a learning rate of $10^{-4}$, while $C_{exp}$ uses a learning rate of $2\times10^{-4}$. We train our framework for $400$ and $40$ epochs on RaFD and MMI datasets respectively, with a batch size of $8$. The learning rate is linearly decayed after half of training epochs. 
The whole framework is jointly trained on a GTX 2080 Ti GPU with about 20 hours.


\begin{figure}[t]
    \centering
    \includegraphics[width=1\linewidth]{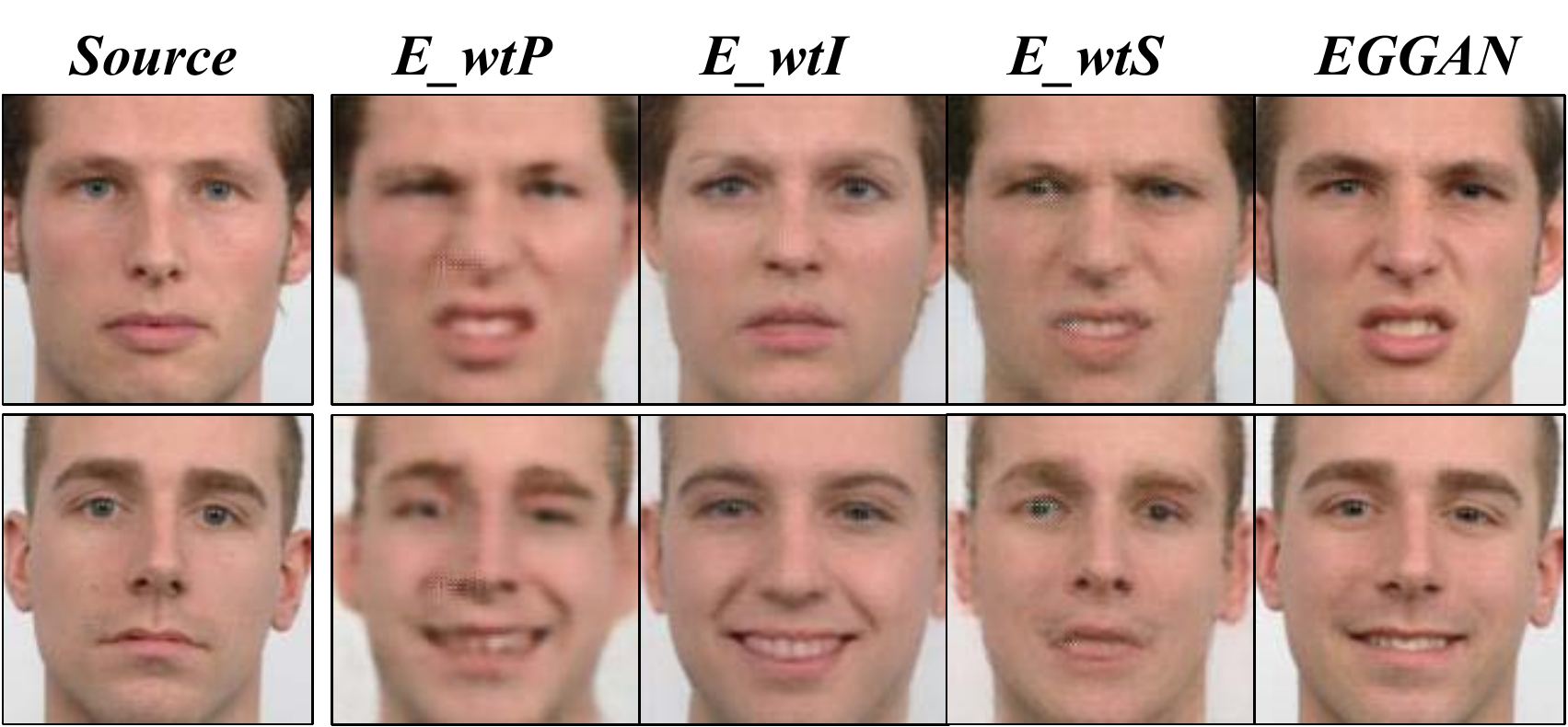}
    \vspace{-5mm}
    \caption{Illustration of the effectiveness of different loss terms. E\_wtP, E\_wtI, and E\_wtS denote training EGGAN without $\mathcal{L}_{per}^{\phi,3}$, $\mathcal{L}_{id}$, and $SSIM(\cdot)$, respectively.}
    \label{fig:ablation}
    \vspace{-0.15in}
\end{figure}

\subsection{Comparison with State-of-the-Art Methods}

\noindent\textbf{Qualitative Results.} We compare our method with HomoInterpGAN~\cite{chen2019Homomorphic} and StarGAN~\cite{Choi_2018_CVPR} on the RaFD dataset, which are two effective and implementable methods in facial attribute manipulation. Since these methods take advantage of discrete attributes and classify expressions into $8$ classes: neutral, angry, contemptuous, disgusted, fearful, happy, sad and surprised, we compare with them for discrete expressions. During the implementation, we treat each expression as a transformation between two domains in StarGAN, and make use of target images to transfer expressions in HomoInterpGAN directly. Fig.~\ref{fig:compairresult} shows the visual comparison of HomoInterpGAN, StarGAN and our method EGGAN on RaFD. We can see that our EGGAN can generate realistic images for all the expressions. StarGAN handles well for image translation in different domains, but its results are a little blurry with some artifacts. HomoInterpGAN gets higher-quality results but cannot keep the identities well.

\noindent\textbf{Quantitative Results.} To quantitatively evaluate the expression manipulation, we compute mean square error (MSE) and Pearson's correlation coefficient (PCC) between the AU intensities of the original target images and generated images. Since we do not have ground-truth for expression manipulation, we employ OpenFace~\cite{baltrusaitis2018openface} to predict the AU intensities. In particular, we sample $1,392$ target images from the RaFD training set with $8$ different expressions and use their AU intensities as the target labels, and we select a source image from the RaFD test set. In this way, we can generate images with expected target expressions. 
Table~\ref{tab:cacuAU} shows the MSE and PCC between 17 AU intensities of target images and synthesized images. 
We can observe that our EGGAN achieves the best results, which demonstrates the effectiveness of inheriting fine-grained expressions from the target images.

To quantitatively evaluate the ability of identity preservation, we conduct face verification by determining whether a source image and a generated image belonging to the same identity. We utilize a released state-of-the-art face recognition model LightCNN~\cite{wu2018light} to obtain the face verification results. Table~\ref{tab:cacuID} show the accuracy of identity preservation. Experimental results show that our method achieves a better performance than HomoInterpGAN and StarGAN.

\subsection{Fine-Grained Expression Interpolation}
Given a target expression label $u_{t}$ and a source image $I_{i}$ with its expression label $u_{i}$, the linear interpolation from source to target expressions can be formulated as
\begin{equation}
    \begin{aligned}
        \mathcal{I}(I_{i},u_{i}, u_{t})=G(l_{i}|u_{i}+\alpha(u_{t}-u_{i}))
    \end{aligned}
\end{equation}
where $\mathcal{I}(I_{i}, u_{i}, u_{t})$ refers to the gradual procedure of manipulating image $I_{i}$, and $\alpha \in [0,1]$ controls the variability of fine-grained expression manipulation. Fig.~\ref{fig:interpshow} shows the interpolation results for example images from RaFD and MMI datasets. We can see that EGGAN can automatically handle fine-grained expression manipulation while preserving identity and global facial shape.

\vspace{-0.1in}
\subsection{Ablation Study}
To further validate each loss term in our framework, we examine how the generated images could degenerate without these loss terms. Let EGGAN be the original model trained with the full objective in Eq.\ref{con:fulllossfunction}. We alternatively train the network by removing the perceptual loss $\mathcal{L}_{per}^{\phi,3}$, identity classification loss $\mathcal{L}_{id}$ and MS-SSIM loss $SSIM(\cdot)$, which are denoted as E\_wtP, E\_wtI and E\_wtS respectively. Fig. \ref{fig:ablation} shows some visual examples. We can see that training without $\mathcal{L}_{per}^{\phi,3}$ leads to the lack of fine semantic information of generated images and some artifacts. The results generated by model trained without $\mathcal{L}_{id}$ cannot keep identities well. Without $SSIM(\cdot)$, the model cannot get effective results with expected expressions. The reason is that the perceptual loss and identity classification loss preserve the high-level semantic information of spatial structural feature and identities, and the MS-SSIM loss helps refine the quality of expression manipulation. 
\vspace{-0.05in}
\section{Conclusion}
We have presented an end-to-end expression-guided generative adversarial network for fine-grained expression manipulation. Moreover, we have incorporated a cGAN with an AAE to learn a structured latent space and generate images with continuous expression labels. Experimental results have demonstrated that our framework achieves a good performance in expression manipulation and can generate realistic images between source and target expressions.

\bibliographystyle{IEEEbib}
\bibliography{reference}

\end{document}